\documentclass{article}

\usepackage{arxiv}

\usepackage[utf8]{inputenc} 
\usepackage[T1]{fontenc}    
\usepackage{hyperref}       
\usepackage{url}            
\usepackage{booktabs}       
\usepackage{amsfonts}       
\usepackage{nicefrac}       
\usepackage{microtype}      
\usepackage{lipsum}		
\usepackage{graphicx}
\usepackage{natbib}
\usepackage{doi}

\title{Making Bielik LLM Reason (Better):\\ A Field Report}

\author{ \href{https://orcid.org/0000-0003-4170-8665}{\includegraphics[scale=0.06]{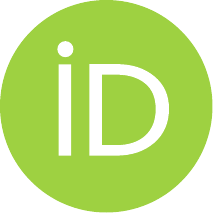}\hspace{1mm}Adam Trybus} \\
	Institute of Philosophy\\
	Jagiellonian University\\
	52 Grodzka St., Kraków 31-044, Poland\\
	\texttt{adam.trybus@uj.edu.pl} \\
	\And
	\hspace{1mm}Bartosz Bartnicki \\
    Bielik.ai (Speakleash Foundation)\\
  33/4 Niemczańska St., Wrocław 50-561, Poland\\
	\texttt{bartosz.bartnicki@speakleash.space} \\
    \AND
	\hspace{1mm}Remigiusz Kinas \\
    Bielik.ai (Speakleash Foundation)\\
  33/4 Niemczańska St., Wrocław 50-561, Poland\\
	\texttt{remigiusz.kinas@bielik.ai} \\
}

\renewcommand{\shorttitle}{Making Belik LLM Reason (Better)}

\hypersetup{
pdftitle={Making Belik LLM Reason (Better): A Field Report},
pdfsubject={cs.CL, cs.AI},
pdfauthor={Adam Trybus, Bartosz Bartnicki, Remigiusz Kinas},
pdfkeywords={Bielik LLM, reasoning models, benchmarking},
}

\begin{document}
\maketitle

\begin{abstract}
	This paper presents a research program dedicated to evaluating and advancing the reasoning capabilities of Bielik, a Polish large language model. The study describes a number of stages of work: initial benchmarking and creation of evaluation methodology, analyzing of comparative results with other LLMs and outlining of future prospects that take into account the limitations of the analyses conducted so far and aims to keep Bielik in the race give the ever-changing --- and competitive --- AI landscape.
\end{abstract}


\section{Introduction}

Frontier artificial intelligence models have begun to outperform humans in programming and mathematical competitions, while systems such as AlphaFold have become indispensable tools in the life sciences. The trajectory of scientific research is increasingly shaped by collaboration between human researchers and AI agents. Significant advances have already been achieved in several areas of physics, highlighting the potential of AI-driven approaches. The overall direction of development points toward a transition from predictive AI systems, through partially assistive tools, to fully autonomous and integrated research platforms orchestrated by AI agents. Some of the major figures in the field, argue that Poland has largely missed the last seven years of global AI progress and remains at the bottom of the EU rankings in AI adoption, just ahead of Romania. The country lacks visibility on the global map of major AI research and technology hubs. While Poland’s AI ecosystem is arguably underdeveloped, there is significant progress to be noted as well. One of the most promising projects is the development of Bielik --- a large language model that can be competitive on the local market with the major international platforms. The model is under continous development (\cite{b1}, \cite{b2}, \cite{b3}, \cite{b4}, \cite{b5}) and has already been incorporated in at least one commercial solution on a big scale.

The reported work is aimed at measuring, comparing and developing the reasoning capabilities of Bielik in order to further increase its competetiveness on the ever-changing AI landscape. We discuss initial research, current efforts, and future prospects. We note that the most recent trend is to go beyond the use of a single model but rather a creation of orchestrated multi-component systems, or pipelines, where LLMs play the role of intermediary among other elements. Those other elements provide functionality unavailable to LLMs, such as strict formal analysis, explainability or repeatability. Our long-horizon goals reflect this trend, also in the context of domain-specific systems e.g. mathematical or legal reasoners.

\section{Initial Efforts}

The work began with the goal of assessing the current reasoning capabilities of Bielik 2.3. The first analyses of reasoning were conducted manually. Its capabilities were compared with the results of state-of-an-art reasoning models. Bielik 2.3 exhibited very limited reasoning abilities — even relatively simple tasks were challenging. Nevertheless Einstein's Riddles drawn from printed sources proved surprisingly effective: they challenged strong models such as DeepSeek-R1 (March 2025), scaled naturally in difficulty through additional parameters and misleading clues, and demanded iterative reconciliation of premises and intermediate conclusions. The principal risk was the "lost-in-the-middle" effect, arising from the considerable length of such tasks.

The three-person research team recognized that manual analysis was inefficient, and thus decided to automate the entire process. As a result, the first automated solution was developed, allowing the addition of tasks and instructions for a metamodel responsible for evaluating the responses. In the first version, ChatGPT-4o served as the “judge,” evaluating both the reasoning process and the final answer. Evaluation was performed on a 1–5 scale, taking into account the correctness and clarity of the response. For the purposes of the project, we outlined a tentative taxonomy of reasoning, and divided the tasks broadly into \textbf{deterministic} (a single well-specified answer) and \textbf{non-deterministic} (requiring interpretation, argumentation analysis, or judgment). In addition, a dozen or so reasoning taxonomies were defined, covering both formal and informal logic, each supplemented with several example tasks.

\textbf{Inductive reasoning tasks}—deducing rules from examples—were indispensable. Sequence continuation (e.g., 2, 6, 18, 54, 162, \dots) or fictional language translation tasks required pattern extraction and rule inference. Induction was a sine qua non condition for general reasoning competence. However, many compelling inductive puzzles were visual rather than textual. Text-only variants often failed to challenge the model meaningfully, although further research might have uncovered stronger candidates. \textbf{Classic riddles} and \textbf{word puzzles} were deterministic and concise, but limited in number and insufficiently demanding in terms of multi-step reasoning. They lacked multi-turn scalability and advanced inferential depth. Non-deterministic reasoning categories introduced additional complexity. \textbf{Paradox analysis}—such as the Barber paradox or the Ship of Theseus—tested conceptual consistency. However, the set of canonical paradoxes was finite, multi-turn potential was limited, and evaluation of solutions was partially subjective. \textbf{Constructing valid arguments} introduced further non-determinism. While multi-turn argument expansion was possible, evaluating the sufficiency and hierarchical structure of premises inevitably required arbitrariness. Attempts to justify justifications risked infinite regress. \textbf{Critiquing argument structures} posed similar challenges. Although models typically performed well in identifying weaknesses in arguments about video games and violence or flawed historical generalizations (e.g., misrepresentations of Einstein’s education), such tasks bordered on writing skill assessment rather than pure reasoning evaluation. Deterministic scoring was difficult. Several additional reasoning categories remained under development: \textbf{syllogistic reasoning, logical fallacy identification, hypothesis formation, and analogical problem solving}. These areas required more systematic task design to ensure objective evaluation criteria.

After much trial-and-error and deliberation, the team settled on question types and desirable answer formats. The final dataset included questions from various sub-categories. Most notably, we tested formal reasoning capabilities with simple tasks in propositional and predicate logic. For example:\footnote{All the questions were originally phrased in Polish.}\\

\begin{footnotesize}

\noindent \textbf{Example 1:} What follows logically from the statements $\lnot p \vee \lnot q$, $p$, and $q \vee r$?\\

\noindent \textbf{Example 2:} Is the statement $\forall x ,\phi \to \exists x ,\phi$ a logical truth?\\
    
\end{footnotesize}

We also designed questions testing various other modes of reasoning, such as abduction or induction. For example:\\

\begin{footnotesize}

\noindent \textbf{Example 3:} Earlier in the morning, you saw several coworkers go into the office kitchen to brew coffee. In the meantime, you had to leave to take care of some errands for your boss. An hour later, while walking down the hallway at work, you pass someone holding a steaming mug with the strong smell of freshly brewed coffee rising from it. You pause for a moment and ask yourself: where did this coffee come from? What conclusion suggests itself? How certain is that conclusion?\\

\noindent \textbf{Example 4:} Assume the following observations are true. Do not make any additional assumptions about the relationships between the listed phenomena. \underline{Observation 1:} Fever (A), cough (B), headache (C), and fatigue (D) occur together with inflammation, dehydration, an immune response, and a viral infection. \underline{Observation 2:} Cough (B), headache (C), and fatigue (D) occur together with inflammation, dehydration, and an immune response.
According to these assumptions, which of the listed phenomena (A–D) is a cause, part of a cause, or an effect of a viral infection? Why?\\

\end{footnotesize}

Moreover, a range of other types of puzzles were incorporated, including queue problems, clock problems, and some trick-answer questions. For example:\\

\begin{footnotesize}

\noindent \textbf{Example 5:} Adam, Betty, and Celine are standing in a line. Adam is not first, Betty is not in the middle, and Celine is not last. What are the possible orders?\\

\noindent \textbf{Example 6:} What time does the digital clock show if it is two minutes to five past half past one?\\

\noindent \textbf{Example 7:} The two halves of the roof on my house are uneven: one slopes at an angle of 60 degrees and the other at 70 degrees. Suppose a rooster lays an egg exactly at the top of that roof. On which side of the roof will the egg fall?\\

\end{footnotesize}

 Within deterministic reasoning, deductive logic puzzles constituted a core benchmark component. These included structured problems requiring reconciliation of explicit constraints: ordering tasks (e.g., determining the sequence of individuals in a queue given relational constraints), rule-based movement puzzles (such as chessboard constraints), and classical “guilty/innocent” scenarios with well-defined premises. Clock puzzles and age-price-time riddles, while superficially logical, tended to collapse into mathematical reasoning and were better classified under math. Full list of taxonomies presented in Table \ref{t1}.

\begin{table}[ht]
\centering
\begin{tabular}{c l c}
\hline
No. & Taxonomy & Deterministic \\
\hline
1  & Abductive reasoning                                    & no  \\
2  & Argument evaluation                                    & no  \\
3  & Argument standardisation                               & no  \\
4  & Classical predicate logic                              & yes \\
5  & Classical propositional logic                          & yes \\
6  & Deductive reasoning (ASCII)                            & yes \\
7  & Deductive reasoning (Einstein Riddles)                 & yes \\
8  & Deductive reasoning (others)                           & yes \\
9  & Detecting contradictions                               & yes \\
10 & Identifying premises and conclusions                   & no  \\
11 & Inductive reasoning                                    & no  \\
12 & Non-classical logics (intuitionism, modal logic)       & yes \\
13 & Nonsense                                               & no  \\
14 & Probability estimation                                 & yes \\
15 & Reasoning by analogy                                   & no  \\
16 & Reasoning by reductio ad absurdum                      & no  \\
17 & Reasoning fallacies                                    & no  \\
18 & Rhetorical devices                                     & no  \\
\hline
\end{tabular}
\caption{Benchmarking Taxonomies}
\label{t1}
\end{table}

With the growing number of questions, we began probing the capabilities of the model. Empirical observations revealed an important limitation. When puzzles were short and tightly constrained, Bielik 2.3 could sometimes correctly identify impossibility or paradox. However, as soon as additional variables were introduced, the model began to hallucinate and sometimes contradicted itself. A particularly destabilizing intervention was modifying the problem statement in a second step: the model struggled to abandon initial assumptions and replace them with new instructions. This rigidity suggested insufficient dynamic belief revision. Despite this, deterministic puzzles had clear advantages: they meaningfully challenged the model at its stage of development and possessed strong multi-turn potential through incremental instruction changes. The drawback was the difficulty of calibrating difficulty levels consistently. Significant progress occurred after a renewed analysis of the results. The evaluation metamodel was replaced with ChatGPT-4.1, which noticeably improved the quality of the assessments. The evaluation system itself was also improved: the new scale used the values 0 / 0.5 / 1, and the separate evaluation of the reasoning process was abandoned in favor of focusing exclusively on the final answer.

\section{Bielik-R Benchmarking}

Meanwhile, the SpeakLeash team released new models: Bielik 2.5 and Bielik 2.6. One of them, after appropriate modifications informed by the initial results of our team, became the basis for the first beta version of Bielik-R: the first Polish large language model equipped with reasoning capabilities. The training pipeline consisted several stages. First, \textbf{supervised fine-tuning} (SFT) was performed using approximately 1.3 million distilled reasoning traces (in English) sourced from several open-source models, including DeepSeek and Qwen. The next stage was \textbf{alignment} by means of Direct Preference Optimization (DPO) using preference data only, without any reasoning-specific training signal. The third stage was \textbf{reinforcement learning} using GRPO (Group Relative Policy Optimization) and DAPO, trained using VERL (Volcano Engine Reinforcement Learning for LLMs) on a set of 143k Polish verifiable tasks. A dedicated reasoning chat template was designed, featuring a modified system message employed during both training and inference; the template introduced explicit \texttt{<think>} and \texttt{</think>} delimiters to demarcate the model's reasoning trace from its final response. Eventually, the team selected 111 of the puzzles and used them in a comparative analysis of various version of Bielik against a selection of other models. The final generated results are presented in Table \ref{t2}.

\begin{table}[ht]
\centering
\begin{tabular}{c l c}
\hline
Rank & Model & Score \\
\hline
1  & gemini-3-pro-preview      & 87\% \\
2  & o3                        & 87\% \\
3  & gemini-2.5-pro            & 84\% \\
4  & grok-4                    & 83\% \\
5  & o4-mini                   & 79\% \\
6  & deepseek-R1               & 78\% \\
7  & o1                        & 78\% \\
8  & seed-oss:36b              & 75\% \\
9  & k2-think:30b              & 71\% \\
10 & qwq:32b                   & 70\% \\
\hline
15 & qwen3:8b                  & 68\% \\
16 & phi4-reasoning:14b        & 68\% \\
17 & magistral-small:24b       & 65\% \\
18 & \textbf{bielik-r:11b}              & 56\% \\
19 & mistral-small3.1:24b      & 45\% \\
20 & \textbf{bielik-v3:11b}             & 45\% \\
21 & \textbf{bielik-v2.6:11b}           & 45\% \\
22 & gemma-3:12b               & 44\% \\
23 & phi4:14b                  & 44\% \\
24 & gemma-3:27b               & 42\% \\
25 & qwen2.5:14b               & 39\% \\
26 & \textbf{bielik-v2.5:11b}           & 29\% \\
27 & \textbf{bielik-v2.3:11b}           & 29\% \\
\hline
\end{tabular}
\caption{Model Performance Ranking}
\label{t2}
\end{table}

We see that all the Bielik models lag behind. This might be due to a variety of reasons, including methodological choices regarding answer checking, as outlined above. What is particularly interesting, however, is how Bielik handles the formal logic tasks. In terms of propositional logic tautology testing, there was only one clear winner gemma-3:12b and answer analysis reveals that the model actually attempted a creation of a truth-table (the standard in-class solution for humans). The question on predicate calculus proved to be suprisingly easy for most models (and predicate calculus is the more complicated logical formalism) --- this does not include the Bielik models, however. This might be an artefact of methodological issues as outlined above or can be due to a relative simplicity of the selected predicate calculus questions. Separately, we also tested the available version of Bielik-R on a full range of formal logic questions and the results are 89\% score in first-order logic and 80\% in propositional calculus. Since then, several programmers have joined the team, which has significantly accelerated the project’s development. The first training sets were created with the goal of broadening the range of topics Bielik-R is familiar with. These sets consist partly of algorithmic tasks automatically generated from existing libraries, and partly of tasks generated by proprietary algorithms and manually prepared problems from classical propositional calculus and first-order logic. For all tasks, an additional Chain-of-Thought reasoning trace from the DeepSeek-R1 model was generated to serve as a training reference for Bielik-R; the underlying datasets were authored by Walery Kusznirski.

Moreover, Bielik-R is also under constant development. Following the release of Bielik-11B-v2.6, subsequent models were trained to reason exclusively in Polish. A hybrid approach was adopted: the model could operate in both reasoning and non-reasoning modes, toggled solely by switching the system message at inference time. To support reinforcement learning with Verifiable Rewards (RLVR), a large-scale Polish dataset was constructed spanning mathematics, code, STEM, and logic domains, comprising 490k tasks, each consisting of a question or task paired with a verified answer. A dedicated data and training pipeline was developed for the Bielik project, encompassing synthetic reasoning data generation, verifiability proofing, and data cleaning. Extensive ablation studies were conducted comparing several RL methods --- GRPO, DR-GRPO, ReMax, and DAPO --- and DAPO was ultimately selected for the production training pipeline. In addition, Polish-language versions of established mathematical benchmarks were created, including translations of AIME, AMC, and MATH-500. Further post-training evaluation employed additional benchmarks such as Olympiad and Minerva.\footnote{We note that the full development of Bielik-R proceeded through three stages. First, an initial proof-of-concept was conducted with Bielik-4.5B-v3.0, trained on 12k MATH samples using 8$\times$ Nvidia H100 GPUs (Athena cluster). Second, Bielik-11B-v2.6 was trained on 143k high-quality Polish samples covering mathematics, STEM, and logic on 32$\times$ Nvidia GH200 GPUs at the Helios Academic Computer Centre (Cyfronet AGH). Third, Bielik-11B-v2.6-R was developed as a dedicated Polish thinking model. This version remains an internal research artifact and has not been publicly released.} The scope of collected metadata is also being actively expanded. The reasoning process has once again been subjected to analysis — now, an additional metamodel evaluates the number of redundant repetitions in the reasoning text.

The comparison between models reveals a structural asymmetry. Gemma, while not particularly strong in reasoning, often receives partial credit even for incorrect answers. Bielik occasionally exceeded the token limit mid-solution, losing points despite being on track to a correct answer — yet its average token consumption remains on par with commercial models (Table \ref{t3}).

\begin{table}[ht]
\centering
\begin{tabular}{rlr}
\hline
\textbf{Rank} & \textbf{Model} & \textbf{Avg.\ Reasoning Tokens} \\
\hline
1 & grok-3-mini             & 1 787 \\
2 & o4-mini                 & 1 800 \\
3 & o3                      & 2 001 \\
4 & o1                      & 2 751 \\
5 & \textbf{bielik-r:11b}            & 3 152 \\
6 & Deepseek-R1             & 3 318 \\
7 & gemini-3-pro-preview    & 3 435 \\
8 & gemini-2.5-pro          & 3 462 \\
9 & grok-4                  & 4 342 \\
\hline
\end{tabular}
\caption{Bielik-R and Commercial Models by Average Reasoning Tokens}
\label{t3}
\end{table}

Additionally, Bielik appears not to “know how to begin” many tasks, indicating gaps in procedural familiarity rather than purely deductive weakness. This observation raises the separate issue of constructing targeted fine-tuning datasets derived from identified weaknesses. There was also a proposal to generate tasks from informal logic in the spirit of Rationale-style reasoning exercises. The operational conclusion was pragmatic: whenever Bielik performs poorly on a specific class of problems, those tasks should be prioritized in the fine-tuning dataset. Approximately 50–100 puzzles should be added in the established format, including more tasks deliberately designed to induce reasoning errors. All such tasks should be represented in ASCII form for consistency.

\section{Future Prospects}

In the next stage, the team plans to broaden the scope of model reasoning research into new, more complex cognitive areas, including discourse analysis — studying coherence, intention, and relationships between utterances in dialogue; Winograd-type tasks — testing the ability to understand context and semantic relationships; and moral dilemmas and conflicts of interest — situations in which the model must balance motivations, consequences, and conflicting reasons or priorities. Future work will extend reasoning research into more complex cognitive domains, including discourse analysis, Winograd-type tasks, and moral dilemmas. A dedicated benchmark for hallucination induction is also planned, with entropy-based metrics integrated into the metadata schema to support comparative evaluation of model robustness.

A deeper theoretical question concerns the boundary between memorization and genuine reasoning: specifically, under what conditions can a model be said to engage in creative inference rather than executing learned algorithms or retrieving stored reasoning paths. Related to this is the observed tendency of models to apply simplifying heuristics — reformulating problems into familiar structures rather than engaging with them in their original form. Identifying the conditions that trigger such behavior, and determining whether it constitutes efficient generalization or undesirable shortcutting, remains an open empirical problem. Collectively, these directions converge on a unified research agenda — distinguishing memorization from inference, characterizing heuristic shortcuts, evaluating cross-domain generalization, and advancing evaluation methodology in a principled and experimentally grounded manner — and naturally structure the future work of the Bielik reasoning project into several coordinated tracks.

\textbf{In formal logic}, the starting point is ensuring the model can reliably execute correct deductive reasoning—essentially mastering valid inference patterns without error. The next phase would involve applying this competence to the interpretation of legal texts, including real-world case studies. A potential avenue of research is minimisation of hallucinations in large language models reasoning about legal issues. \textbf{In informal logic}, the initial step is to familiarize the model with argumentation trees, eristic techniques, and related analytical tools. Building on that foundation, the next objective would be to develop a “Demagog”-style system capable of analyzing and evaluating extended argumentative discourse. Regarding \textbf{mathematics}, the first step is consultation to determine a realistic and well-defined scope of mathematical competence for the model. From there, the project should either identify a specific mathematical niche to specialize in or conduct a rigorous evaluation of the model’s capabilities at a clearly defined and practically useful level. Bartosz Naskręcki, PhD — author of a mathematical benchmark no language model has yet solved — joined the analytical team and directly shaped the development of \textbf{Bielik-M}, a multi-agent system for solving Polish matura mathematics problems powered by Bielik 3, built by Patryk Orwat. System pipeline comprises four specialized agents — Analytical (method identification), Executor (SymPy code generation with auto-sanitization and self-repair), Summary (pedagogical step-by-step explanation with exam review topics), and an optional Lean 4 Formalizer — augmented by a local RAG service (TF-IDF with Polish synonyms) indexing 53 math methods and 200+ historical exam problems. The key finding is that even a relatively small 11B model can effectively solve exam-level math when supported by proper task decomposition into specialized agents, symbolic verification (SymPy), and contextual retrieval. This paves the way for the creation of an AI tutor in Polish. Finally, in the area of \textbf{games and puzzles}, we focus on training models to discover winning strategies in board and video games through \textbf{multi-turn interaction}. We began with experiments in which benchmarked models were required to iteratively improve their strategy in turn-based games, within a framework developed and actively maintained by Krystian Kulas, with Model Context Protocol (MCP) integration enabling standardized communication between the model and the game environment. An example result from a game played by Claude Opus 4.5, shown in Table \ref{t4}, demonstrates that the model needed seven battles to develop a winning strategy.

\begin{table}[ht]
\centering
\begin{tabular}{|c|c|c|}
\hline
\textbf{Battle} & \textbf{Strategy} & \textbf{Result} \\
\hline
\#1--2 & Learning basics & Partial success \\
\#3    & Sequential elimination & 3 eliminations \\
\#4    & Aggressive Melee & 4 eliminations (so close) \\
\#5    & Shoot enemy Rangers & 1 elimination \\
\#6    & Defensive Melee & 0 eliminations (failure) \\
\#7    & Return to 4. style & victory \\
\hline
\end{tabular}
\caption{Battle Results Summary Generated by Claude Opus 4.5}
\label{t4}
\end{table}

\section{Conclusions}

In this paper, we described a research program focused on evaluating and improving the reasoning capabilities of Bielik, a Polish large language model intended to strengthen Poland’s position in the AI landscape. While Poland has lagged behind global leaders in AI adoption, Bielik is presented as a promising local initiative. The project aims not only to benchmark Bielik’s reasoning performance but also to develop it toward integration within multi-component AI systems, where LLMs function as intermediaries in larger, orchestrated pipelines.

In many ways, the work on improving the reasoning capabilities of Bielik has just begun. So far we have compared how various versions on Bielik fared in comparison to other top-notch LLMs on a selection of tasks and deeply tested Bielik-R on a full range of formal reasoning tests prepared by the team. The conclusions are mixed (also due to the limiting factors outlined above): while Bielik still lags behind on many tasks, there are those where the model is decidedly mid-pack or even getting ahead of the competition (as the reasoning difficulty analysis seems to reveal). No doubt more work is required, including additional model training. The immediate next steps relate to continued work in the directions outlined but also a move in other directions, probing continual learning and domain-specific skills of various incarnations of Bielik. Moreover, we plan to focus on multi-component agentic domain-specific systems (in mathematics, law or critical thinking) that would incorporate Bielik as one of the components.

\bibliographystyle{unsrtnat}
\bibliography{references}  






\end{document}